\pdfoutput=1
\documentclass[11pt]{article}

\usepackage{authblk}
\usepackage[]{ACL}

\usepackage{algorithm}
\usepackage{amsmath}
\usepackage{algorithmic}

\usepackage{microtype}
\usepackage{array}
\usepackage{enumitem}
\usepackage{subfig} 
 
\usepackage{enumitem}
\usepackage{subfig} 

\usepackage{times}
\usepackage{latexsym}

\usepackage[T1]{fontenc}
\usepackage[utf8]{inputenc}

\usepackage{microtype}

\usepackage{inconsolata}

\usepackage{graphicx}

\usepackage{times}
\usepackage{latexsym}

\usepackage[T1]{fontenc}
\usepackage[utf8]{inputenc}

\usepackage{microtype}
\usepackage{inconsolata}

\title{Improving Grounded Language Understanding in a Collaborative Environment by Interacting with Agents Through Help Feedback}

\makeatletter
\newcommand\email[2][]%
   {\newaffiltrue\let\AB@blk@and\AB@pand
      \if\relax#1\relax\def\AB@note{\AB@thenote}\else\def\AB@note{\relax}%
        \setcounter{Maxaffil}{0}\fi
      \begingroup
        \let\protect\@unexpandable@protect
        \def\thanks{\protect\thanks}\def\footnote{\protect\footnote}%
        \@temptokena=\expandafter{\AB@authors}%
        {\def\\{\protect\\\protect\Affilfont}\xdef\AB@temp{#2}}%
         \xdef\AB@authors{\the\@temptokena\AB@las\AB@au@str
         \protect\\[\affilsep]\protect\Affilfont\AB@temp}%
         \gdef\AB@las{}\gdef\AB@au@str{}%
        {\def\\{, \ignorespaces}\xdef\AB@temp{#2}}%
        \@temptokena=\expandafter{\AB@affillist}%
        \xdef\AB@affillist{\the\@temptokena \AB@affilsep
          \AB@affilnote{}\protect\Affilfont\AB@temp}%
      \endgroup
       \let\AB@affilsep\AB@affilsepx
}
\makeatother

\usepackage{lipsum}

\author{\textbf{Nikhil Mehta}$^{1}$\thanks{\phantom{\_\_}Work done during an internship at Microsoft Research.} \quad \textbf{Milagro Teruel}$^2$ \quad \textbf{Patricio Figueroa Sanz}$^3$ \quad \textbf{Xin Deng}$^3$ \\
\textbf{Ahmed Hassan Awadallah$^3$ \quad Julia Kiseleva$^3$} \\
$^1$Purdue University \\
$^2$Universidad Nacional de Córdoba \\
$^3$Microsoft \\
\texttt{mehta52@purdue.edu}}

\begin{document}
\maketitle

\begin{abstract}
Many approaches to Natural Language Processing tasks often treat them as single-step problems, where an agent receives an instruction, executes it, and is evaluated based on the final outcome. However, language is inherently interactive, as evidenced by the back-and-forth nature of human conversations. In light of this, we posit that human-AI collaboration should also be interactive, with humans monitoring the work of AI agents and providing feedback that the agent can understand and utilize. Further, the AI agent should be able to detect when it needs additional information and proactively ask for help. Enabling this scenario would lead to more natural, efficient, and engaging human-AI collaboration.
In this paper, we investigate these directions using the challenging task established by the IGLU competition, an interactive grounded language understanding task in a MineCraft-like world. We  delve into multiple types of help players can give to the AI to guide it and analyze the impact of this help on behavior, resulting in performance improvements and an end-to-end interactive system.
\end{abstract}

\section{Introduction}
\label{sec:intro}
One of the long-lasting goals of AI agents~\cite{winograd1972understanding} is the ability to seamlessly interact with humans to assist in solving tasks. To achieve this, the agent must be able to understand human language and respond to it, so it can execute instructions~\cite{ skrynnik2022learning} or ask clarifying questions~\cite{aliannejadi-etal-2021-building}. Researchers have proposed a large number of tasks aimed at tackling this human-AI collaboration challenge, many based on humans providing instructions to the agent to solve a goal~\cite{gluck2018interactive, shridhar2020alfred}. An example is the blocks world task, where the agent understands human instructions to move blocks on a grid~\cite{bisk2016natural}.

\begin{figure}[t!]
  \centering
  \includegraphics[scale=0.35]{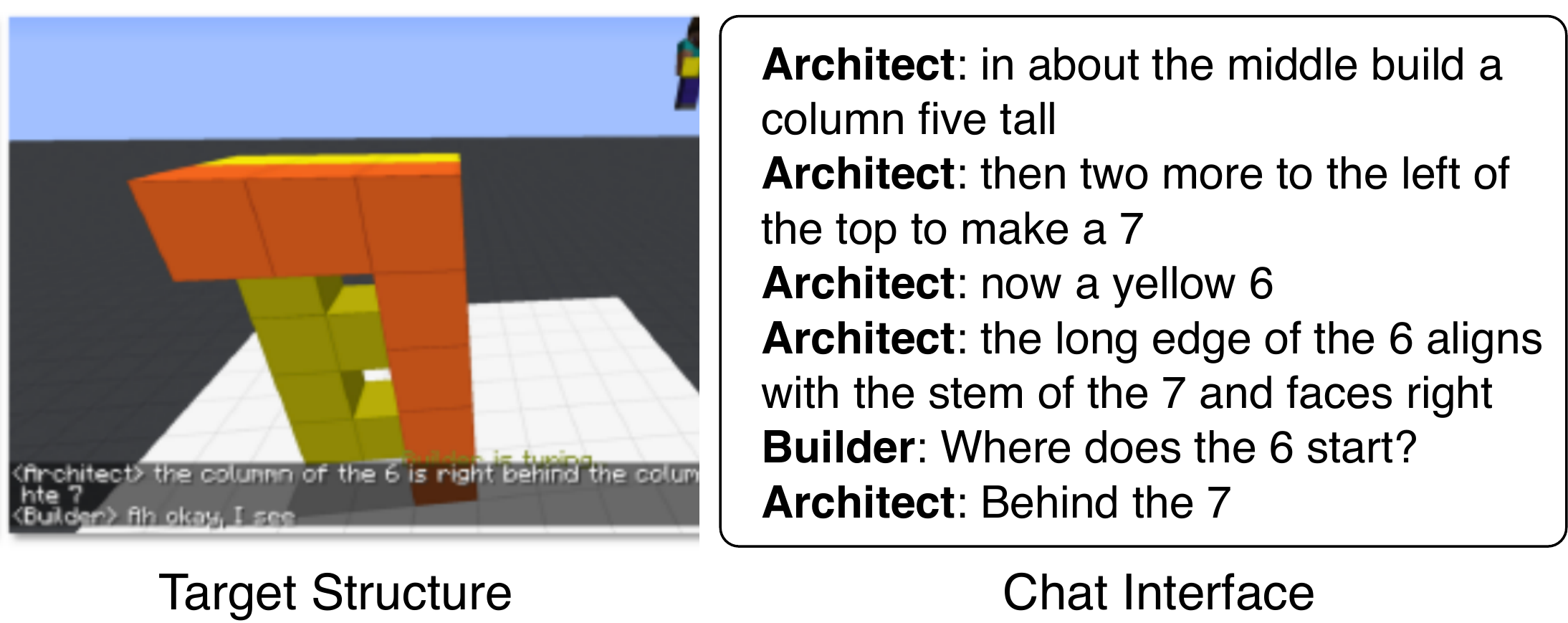}
  \caption{\small An example of the building IGLU task, collected using all human data: Based on the Target Structure (left), the Human Architect provides instructions to the Builder via the Chat Interface (right). As shown, during data collection the human Builder also responds.}
  \label{fig:exampleIGLU}
  \vspace{-15pt}
\end{figure}

A more recently proposed human-AI instruction-based interaction task, is Interactive Grounded Language Understanding
in a Collaborative Environment (IGLU)~\cite{mohanty2023transforming}, where agents collaborate with humans to build a reference structure in the MineCraft 3D world, by placing blocks on a grid. Fig.~\ref{fig:exampleIGLU} illustrates the building task, where the human \textit{Architect}~\cite{narayan-chen-etal-2019-collaborative,jayannavar2020learning} provides instructions to the AI \textit{Builder} agent, via a Chat Interface, to build the \textit{Target Structure}. The IGLU task is particularly challenging since human architect instructions are complex, often referring to broad spatial concepts in the 3D world, such as ``in about the middle build a column five tall''. Understanding these concepts and executing the instructions successfully, even for state-of-the-art systems, is challenging and well below human performance~\cite{kiseleva2022iglu}. 

\begin{figure*}[t!]
  \centering
  \includegraphics[scale=0.33]{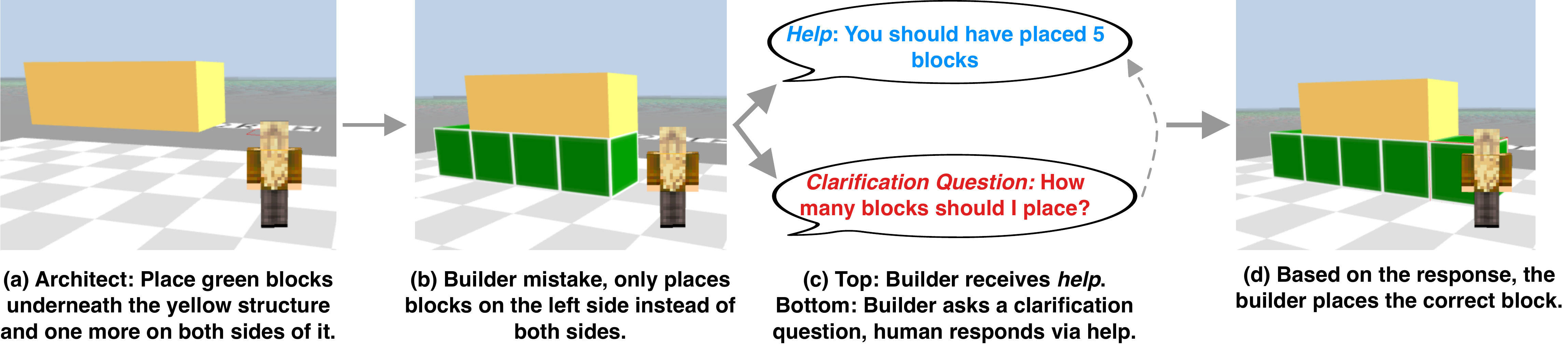}
\caption{\small Our framework overview: \textbf{Improving Grounded Language Understanding in a Collaborative Environment by Interacting with Agents using Help Feedback:} Based on the initial architect instruction (a), the Builder Agent places blocks (b). Noticing a mistake has been made, the human can \textit{interact} to provide \textit{help} (c top), in this case telling the model how many blocks to place. This easy to provide help enables the Builder to solve the task better, leading to a correct prediction (d). Further, the Builder can self-detect confusion and realize it may make a mistake, asking a Clarifying Question (c bottom), which the human can respond to via help (c top), leading to a better prediction (d).}
\label{fig:framework_overview}
\vspace{-15pt}
\end{figure*}

Typically, tasks such as IGLU are evaluated single-step, where an agent is given an instruction, executes it, and is evaluated to obtain final results. However, language is inherently \textit{interactive}, where humans converse back and forth with each other. In this paper, inspired by previous work~\cite{mehta2019improving}, we adopt a different approach, and propose multiple ways in which the AI agent can interact with humans to solve the IGLU task. Specifically, we propose ways in which humans can interact with AI agents to correct their mistakes, by offering \textbf{four different forms of \textit{help}}, a form of online feedback. Following \citeauthor{mehta2019improving}, we define \textit{help} as a high level feedback to the model, that allows it to solve the current task better and learn knowledge for the future. For example, after the agent makes a mistake and places too many blocks on the grid, one form of \textit{help} informs the agent how many blocks it should have placed. While not solving the task directly, this \textit{help} makes the task easier, which has multiple benefits: 
    \textbf{(1)} \textit{Help} enables the agent to make a better prediction on the current instruction (i.e. the model knows how many blocks to place). \textbf{(2)} \textit{Help} provided at training enables the agent to learn better for the future. For example, once the agent knows how many blocks to place, it can focus on learning other aspects of the instruction (such as where to place the blocks), which can generalize to future instructions, where similar concepts may apply. \textbf{(3)} \textit{Help} is simple for humans to provide, as humans don't need to solve the final task, allowing humans to interact with agents easily.

Each form of help we propose is based on a high-level \textit{concept} that is useful for the IGLU task. Through it, the agent is able to understand and take advantage of interactions from humans beyond the initial instruction, to do better. However, in a true interactive scenario the agent should also be able to speak to the human, even unprovoked. To enable this, we propose a method based on \textit{help} in which the agent can self-identify confusion, and use it to ask an appropriate clarification question to the human. This is done by the agent first providing itself several different forms of help (which needs no human interaction and can be done using a separate ML model) until it identifies a concept it doesn't understand. Then, it asks a clarification question based on that concept. Combined with understanding and following help from above, this enables the agent to be fully interactive. It can detect when it's confused, ask for help, and then utilize that help effectively. Experiments show performance improvements. Fig.~\ref{fig:framework_overview} shows an overview.% of our framework.

In summary, we make the following contributions: 
\textbf{C1:} A framework to tackle tasks like IGLU in an interactive manner, where human Architects can have a back and forth interaction with AI agents.
\textbf{C2:} Four different forms of \emph{help}, based on relevant IGLU concepts, that humans can use to help AI agents, specifically when they make mistakes. %(Sec.~\ref{sec:models}) 
\textbf{C3:} A method for agents to self-generate this help, so human interaction is not necessary.
\textbf{C4:} A novel method to take advantage of help for the agent to detect when it's confused, and ask a relevant clarification question.
\textbf{C5:} Performance improvements in these settings, enabling a true interactive agent for solving tasks like IGLU. 

Sec.~\ref{sec:task_specific_models} describes our basline, Sec.~\ref{sec:models} discusses the help we propose and how we use it. Finally, Sec.~\ref{sec:experiments} presents results, and Sec.~\ref{sec:discussion} analyzes them.

\section{Related Work}
\label{sec:related_work}
\noindent
\textbf{Human-AI Interaction Tasks} The task of humans interacting with AI agents to solve real-world tasks is a long-standing problem \cite{winograd1972understanding, clark1996using, koller2010report, narayan2017towards, padmakumar2022teach}. Among other challenges, the embodied AI agent needs to understand complex human language~\cite{kiseleva2016understanding}, spatial world orientation, and unseen concepts~\cite{wang2023voyager}. As this problem is still challenging, datasets like IGLU~\cite{kiseleva2021neurips, mohanty2023transforming}, BASALT~\cite{shah2021minerl,milani2023bedd} and MineDojo~\cite{fanminedojo} have been recently proposed. In this work, we focus on building an agent that understands instructions to place blocks on a grid.

\noindent
\textbf{IGLU Task} Since the IGLU task was proposed \cite{kiseleva2021neurips,kiseleva2022iglu, mohanty2022collecting}, it has been the subject of multiple competitions, such as a RL task (building a RL-based first-person agent to place blocks)~\cite{skrynnik2022learning,zholus2022iglu} and a NLP task (determining when and what clarification questions to ask)~\cite{mohanty2023transforming}. In contrast, as we are interested in building a fully interactive agent, we focus on a \emph{dialogue only IGLU task setup}, where an instruction is provided and a model predicts the blocks to be placed. As we do not focus on building a RL agent and we do not use a vision component (the 3D world space is encoded as language in our setup), our work is not directly comparable to the existing IGLU baselines. However, we use similar metrics when applicable. Further, we hypothesize that our interactive framework can be applied to other IGLU-based tasks, by adding a language component that understands help similar to this paper, and leave it for future work. 

\noindent
\textbf{User-Feedback} As tasks like IGLU are difficult, a crucial component of human-AI interactive systems is the ability of the agent to receive direct feedback from humans, to improve performance. This has been studied in active learning \cite{ren2021survey}, LLM feedback \cite{madaan2023self, akyurek-etal-2023-rl4f}, robotics \cite{ren2023robots}, summarization \cite{shapira2021extending}, and others (see Appendix~\ref{appendix:additional_related_works}). Closest to us, \citeauthor{mehta2019improving} show how hints can be provided to the model. We build upon their regional (``top right'') and directional (``move left'') hints, to enable more forms of user feedback, by proposing additional types of hints. Further, compared to \citeauthor{mehta2019improving}, we evaluate on a significantly more challenging task and use a stronger baseline model (LLMs). We also propose a novel approach for the agent to identify when it is confused, and then enable it to ask relevant clarification questions.

\noindent
\textbf{Clarifying Questions} As instructions may be vague or unclear, the AI agent should be able to ask clarification questions~\cite{aliannejadi2020convai3, aliannejadi-etal-2021-building,arabzadeh2022unsupervised}, to solve the task better. This is often studied, especially in dialogue systems, and is still challenging~\cite{white2021open, kim2021deciding, shi2022learning, manggala2023aligning}. We use our ``help'' to determine when the model is confused and should ask a clarification question, and the question is based on what ``help'' the model needs.

\section{Task-Specific Models}
\label{sec:task_specific_models}
In this section, we first discuss the specific formulation of IGLU we use, which is different from other IGLU setups~\cite{kiseleva2022iglu}, and unique to us. We then briefly explain the model we use for it.

\noindent
\textbf{Task Formulation:} The IGLU task~\cite{kiseleva2021neurips} involves two players, a Builder and an Architect, that collaborate to build a target structure in the 3-D Minecraft world (Fig.~\ref{fig:exampleIGLU}). The Builder places blocks based on Architect's instructions. In our version of this task, the Architect is a human, while the Builder is an AI agent. Thus, the Builder places blocks and subsequently makes mistakes/needs to ask for clarification, while the human Architect (which we simulate) helps the Builder. Further, our task formulation is \emph{fully language-based}, and there is no vision component. This is because we are primarily interested in how to make agents more interactive, and interactions typically happen via language. Hence, we chose this setup for simplicity. Thus, our task formulation is as follows: \emph{Given the Architect and Builder history complete with the last instruction, and a dialogue representation of the current Minecraft World State, predict the coordinate locations of the blocks to place.} 

\noindent
\textbf{Model Architecture} We now briefly discuss the baseline system we train for the Builder model, based on \citeauthor{zholus2022iglu}. Later, we will incorporate help into this model. Our baseline model is a standard BART-base Transformer~\cite{lewis2019bart}, trained for Conditional Generation using the Hugging Face package~\cite{wolf-etal-2020-transformers}. As this is a language model, all of its' inputs and outputs are in natural language. Thus, we now discuss the method we used to convert $(x, y, z)$ coordinate block locations in the Minecraft World Grid to language, so they can be passed into the model as textual input. We first determine how far the coordinate is from the origin of the grid $(0, 0, 0)$, for each axis. In language, we define the x-axis as `left/right', y-axis as `up/down', and z-axis as `higher/lower'. We then combine the distance and direction into a sentence, e.g. an x location of $-2$ would be ``\textit{2 left}'' and a z location of $3$ would be ``\textit{3 higher}''. We ignore model outputs that do not follow this format, as they are invalid. Input grids with multiple blocks can also be encoded into language the same way, just with multiple sentences such as ``\textit{2 left 1 up 3 higher. 4 right 2 down 4 lower.}''.

\section{Help-Specific Models}
\label{sec:models}
While the Builder model introduced in Sec.~\ref{sec:task_specific_models} achieves competitive IGLU performance (Sec.~\ref{sec:help_results}), it still makes a large number of mistakes. Thus, in this paper, we propose an interactive setup, where a human can interact to ``\textit{help}'' the model when it makes a mistake. Rather than telling the model where to place the blocks, which would be difficult to provide and learn from, we propose that humans ``\textit{help}'' the model by assisting it with a high-level concept necessary to solve the final task, making it easier. While not only being simpler to provide than solving the final task, this ``\textit{help}'' enables the model to learn the task better, to perform better when no help is provided (it can focus on other aspects of the task, different from the concept provided by the help; for results see Sec~\ref{sec:help_results}). For example, through one form of help, ``\textit{length help}'', humans assist the model by telling it how many blocks to place. Once the model understands this, it can focus and better learn other aspects of the task instruction, such as where actually to place each block. In this paper, we experiment with humans providing help via a natural language sentence.

We first introduce 4 different forms of help feedback humans can provide agents, all based on different high-level concepts relevant to the IGLU task (Sec.~\ref{sec:subsection_help_types}). Two were introduced by \citeauthor{mehta2019improving}, and others are novel to this work. Detailed ex. of help are in App.~\ref{appendix:help_examples}. Then, in Sec.~\ref{sec:incorp_help}, we discuss how this help can be learned and effectively incorporated into the task-specific baseline from Sec.~\ref{sec:task_specific_models}~\cite{raffel2020exploring}. Finally, in Sec.~\ref{sec:clarifying_questions}, we explain how agents can leverage their comprehension of various forms of help to aid their own performance, effectively identifying when they are confused and then asking clarification questions. This final step culminates in a genuine interactive scenario, where the agent can receive interactions in the form of help and reciprocate by seeking clarifications. Notably, when agents help themselves, they can exhibit improved performance \textbf{without requiring any human interactions.}

\subsection{Help Types}
\label{sec:subsection_help_types}

\textbf{Restrictive:} Similar to \citeauthor{mehta2019improving}, restrictive help restricts the search space of the agent to a general region, such as \textit{top left} or \textit{lower right}. The regions are determined by dividing the grid based on the number of regions desired and then choosing the appropriate one based on the true block location (if multiple blocks are placed by a single instruction, we choose the region randomly from the set of valid ones). Restrictive help significantly simplifies the challenging task of determining where to place blocks, allowing the agent to perform better and learn better for the future when it is provided. An example: \textit{``Place the block in the top left region.''}. We experimented with two ways of forming the regions. The first divides the grid equally, leading to 4 regions total. The second divides the center equally (center divided into 4 or 8 regions) and then the rest of the grid equally (divided into 4 regions) for a total of 8 or 12 regions (4 or 8 from the center and 4 from the non-center).

\begin{figure*}[t!]
  \centering
  \includegraphics[scale=0.26]{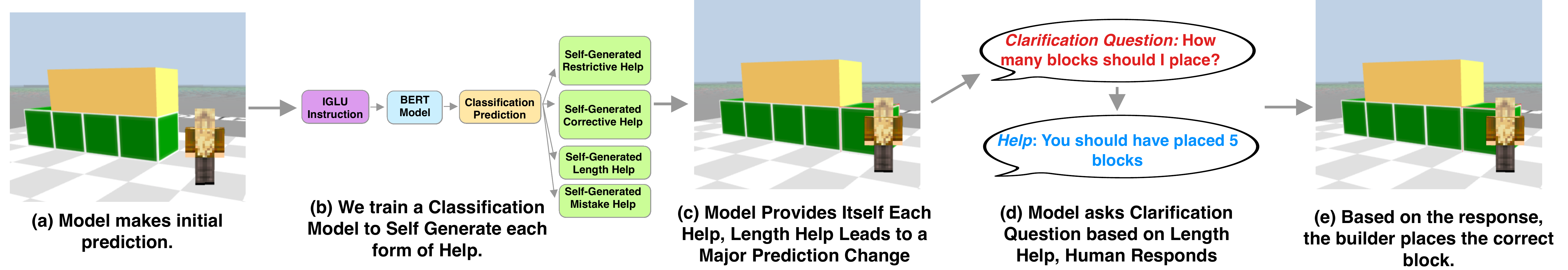}
\caption{\small Framework to Detect Confusion and Ask Clarification Questions: After a model's initial prediction (a), we train a separate classification model (b) to self-generate each form of help. The model takes in the IGLU Architect instruction, and trains a BART Model to predict the appropriate help. For example, for length help, it predicts how many blocks to place (0-6). The agent then provides itself each help, and determines if any help leads to a significant prediction change (c). If it does, the model detects that it is confused and asks a clarification question based on that help (d). Based on the response, the builder places the correct block (e).}
\label{fig:clarification_framework_overview}
\vspace{-15pt}
\end{figure*}

\noindent
\textbf{Length-based:} Length-based help informs the agent how many blocks to place, and if they should be placed together, e.g. a tower. This help is especially useful for instructions involving length-based keywords. Ex: ``\textit{You should place 3 blocks.}''.

\noindent
\textbf{Corrective:} Also similar to \citeauthor{mehta2019improving}, corrective help is provided after observing the agent's initial prediction, and then determining which direction (\textit{up, down, left, right}) to adjust it so it is closer to the target. This enables the agent to improve on its prediction while also restricting the search space by one direction (like the agent only having to look `\textit{left}'). Ex: \textit{``Look left''}

\noindent
\textbf{Mistake-based:} Mistake-based help is also provided after the agent's initial prediction. However, rather than adjusting the prediction's direction, mistake-based help is count-based. Specifically, it makes it easier for the agent to recover from mistakes, by telling it exactly how many blocks it placed incorrectly. Ex: \textit{``2 blocks are wrong.''}

\subsection{Forming Help in Language}
\label{sec:slot_filling_help}
To generate help utterances without having humans provide it (which would be costly), we use synthetic utterances, generated via slot filling. We have utterances with placeholders such as \textit{Place the block in the \_ region} (for restrictive help) and then the slot can be filled in with the appropriate region based on where the block should be placed (which can be determined based on the gold data). We use different language at train and test time, to simulate real humans.  Detailed examples in App.~\ref{appendix:help_examples}. To account for even more language variety than pre-defined utterances, we also use LLMs to simulate real humans providing help in Sec.~\ref{sec:discussion}. 

\subsection{Incorporating Help}
\label{sec:incorp_help}
Incorporating help into the task-specific Builder model from Sec.~\ref{sec:task_specific_models} is important as accurately understanding help is a critical part of being able to use it effectively. To successfully do this, we provide the help as an input to the BART dialogue model, appended as a natural language sentence to the end of the IGLU instruction. For example, the input to BART could be: \textit{``INSTRUCTION:..., HELP: ...''}

We additionally experimented with pre-training a model to learn help as in \citeauthor{mehta2019improving}, but that led to worse results as BART couldn't successfully incorporate the pre-trained layers.

\subsection{Using Help for Clarifying Questions}
\label{sec:clarifying_questions}
In addition to receiving interactions from humans, end-to-end interactive agents should be able to communicate with humans, even unprovoked. One way to do this, which we explore in this paper, is for the agent to self-identify confusion and ask intelligent clarifying questions, when confused. This is particularly important as without it, agents would make predictions even when they are confused, and thus those predictions are likely to be incorrect. Further, agents that ask intelligent questions to humans are more likely to receive better responses than ones that don't, and thus will perform better, particularly if they can understand the responses.

Inspired by these ideas, in this paper, we propose to use \textit{help} feedback to identify confusion and ask clarification questions. First, we focus on identifying confusion. We hypothesize that an agent is confused if it \emph{significantly changes its predictions after receiving help}, as this means the help greatly benefited/hurt the initial prediction. Thus, the agent likely didn't understand the initial instruction well, and was probably confused by it. In this case, we believe the agent should ask a clarification question, based on the concept (or \textit{help} type) that caused the significant prediction change, to avoid making an incorrect prediction. 

As the agent must identify confusion itself, it cannot receive \textit{help} from humans. However, based on our methodology, the agent determines that its confused if its predictions change significantly after receiving \textit{help}. Thus, the agent needs to be able to \textbf{provide itself \textit{help}}, make predictions based on that \textit{help}, and then ask clarifying questions. 

To enable agents to provide themselves help, which is an interesting task, we are inspired by \citeauthor{mehta2019improving}, who propose model self-generated advice, which is a way to generate help \textbf{without human intervention}. The broad idea is to build a classification model and train it to predict the help the agent needs to provide itself. For example, for restrictive help and the IGLU task, the model takes in the IGLU specific dialogue input and predicts what region to place the block in (4 regions $\rightarrow$ 4 way classification problem). Then, based on the region predicted, we can automatically generate the help. For example, if the model predicted region 3, the top left region, the generated help sentence would be: \textit{`Place the block in the top left''}. Intuitively, in this self-generated help setup, the agent is solving a simpler classification task first, using that to generate help, and providing that as input for the more complicated final task of placing blocks. We explain more details of our self-generated help models, including classification objectives for each help type, in Sec.~\ref{sec:self_gen_help_model}.

Once the agent is able to self-generate all forms of help discussed in Sec.~\ref{sec:subsection_help_types}, it can provide itself all of them iteratively, and see where its output prediction changes the most compared to the model that doesn't receive any help (we look at the number of blocks placed). If it is over a threshold (i.e. the number of blocks placed by the agent with self-generated help is significantly more than the agent without the help), we hypothesize that the agent is confused. Then, for that help, the agent can ask a clarification question based on the help, such as ``\textit{What quadrant should the block be placed in?}'' if restrictive help was chosen, and the human can respond by providing help, as in Sec~\ref{sec:subsection_help_types}. Assuming we have learned the model to incorporate help from Sec~\ref{sec:incorp_help}, the agent will be able to understand the human help and take advantage of it for the final IGLU block prediction task. Below, we discuss the models we use to self-generate the help. Algorithm~\ref{alg:confusion_clarification_questions} and Fig.~\ref{fig:clarification_framework_overview} detail the above process for how an agent can take advantage of self-generated help to detect confusion and ask clarification questions.

\begin{algorithm}

\caption{\textit{Detecting Confusion and Asking Clarification Questions} \newline \small Overview: The IGLU task model first generates an initial prediction without help. Then, we iterate through all forms of help, self-generating and feeding them into the IGLU model. The help that leads to the biggest difference in model prediction, if it is bigger than a hyper-parameter threshold, is used to generate a clarification question. The clarification question list is pre-defined and slot-filled based on the help chosen.}
\begin{algorithmic}[1]
  \small
    \STATE \textbf{Input:} $D$ (IGLU Architect Dialogue), $G$ (Current Grid State), $H$ (All Help Types)
  \STATE \textbf{Output:} $Q_s$ (clarification questions)

\STATE $o_{0} = m(d_0, g_0)$ {\scriptsize{\fontfamily{pcr}\selectfont Run IGLU Model}}
  \STATE $o_m = 0$ {\scriptsize{\fontfamily{pcr}\selectfont PlaceHolder for Max Difference from Initial Prediction}} \\
  \STATE $h_m = 0$ {\scriptsize{\fontfamily{pcr}\selectfont PlaceHolder for Most Impactful Help}} \\

% \INPUT  
\FORALL[loop over all help]{$i=1, \dots, n$}
  \STATE $h_i = f_{h_i}(d_i, g_i)$ {\scriptsize{\fontfamily{pcr}\selectfont Generate Help}} \\
  \STATE $o_i = m(d_i, g_i, h_i)$ {\scriptsize{\fontfamily{pcr}\selectfont Run IGLU Model with Help}} \\
  
  \IF[If Difference to Initial is More than Max Difference So Far]{$o_i - o_0 > o_m$}
    \STATE $o_m = o_i$ {\scriptsize{\fontfamily{pcr}\selectfont Store as New Output}}
    \STATE $h_m = i$ {\scriptsize{\fontfamily{pcr}\selectfont Store as Max Help}} \\
  \ENDIF
\ENDFOR

\IF[Max Different Below Threshold No Clarification Question]{$o_m < $ threshold}
    \RETURN 0
 \ENDIF

 \STATE \textbf{Choose Question} $q = q_m(h_m) $
    {\scriptsize{\fontfamily{pcr}\selectfont
Choose Clarification Question From Help}}

\RETURN $q$ (Clarification Question)

 \end{algorithmic}
\label{alg:confusion_clarification_questions}
\end{algorithm}

\subsubsection{Self-Generated Help Models}
\label{sec:self_gen_help_model}
We now discuss more details of the self-generated help models, which are used to generate help that the model provides to itself to determine confusion and generate clarification questions. The self-generated help models are classification based BART-base models. As in the BART for conditional generation model used for the IGLU Builder Task in Sec.~\ref{sec:task_specific_models}, they take in the architect history complete with the last instruction, and a dialogue representation of the current grid. Below, we detail the specific classification goal of each model:

\noindent
\textbf{Restrictive Help} The model is trained to output one of the regions the block must be placed in.

\noindent
\textbf{Length-Based} The model is trained to predict one of 7 classes, corresponding to how many blocks must be placed. There are 6 classes referring to 0-5 blocks, and the 7th refers to more than 5 blocks.

\noindent
\textbf{Corrective Help} The model must output one of 4 directions the predictions must be adjusted towards. In addition to the original input, this model also takes in a grid with the blocks placed based on the most recent Architect instruction, as that is what it needs to adjust its prediction based off of.

\noindent
\textbf{Mistake-Based} The model learns how many blocks must be adjusted. There are 7 classes, corresponding to how many blocks must be adjusted, and None. This model also takes in an additional input grid with the blocks placed based on the most recent Architect instruction.

\section{Experiments}
\label{sec:experiments}

\begin{table*}
\begin{center}
\begin{tabular}{|p{5.2cm}|p{2.4cm}|p{2.1cm}|p{2.1cm}|p{1.8cm}|}
  \hline
  {\textbf{Model}} & {\textbf{ Distance}}
  & {\textbf{Reward}}  & {\textbf{\# Blocks Placed}} & {\textbf{\% Help Followed}} \\

 \hline
 \small $\mathtt{M1:}$ BART Language Model & \small 12.64 (51.75) & \small 1.26 (1.49) & \small 2.56 (2.10) & \small 86.78 (33.86) \\
 \hline 
 \small $\mathtt{M2:}$ Restrictive Help Model Add. Input & \small \textbf{11.64 (53.01)} & \small 1.39 (1.60) & \small 2.84 (2.32) & \small 84.48 (36.20) \\
 % add a stats table w/ # of blocks / region in the gold data
 \small $\mathtt{M3:}$ Correct Help Model Add. Input & \small 11.66 (58.17) & \small \textbf{1.66 (1.85)} & \small 2.80 (2.35) & \small 61.78 (48.59) \\
 \small $\mathtt{M4:}$ Length Help Model Add. Input & \small 18.93 (48.03) & \small 1.32 (1.61) & \small 2.80 (2.47) & \small 61.31 (48.70) \\
 \small $\mathtt{M5:}$ Mistake Help Model Add. Input & \small 16.18 (78.84) & \small 1.46 (1.62) & \small 2.68 (2.28) & \small 93.24 (25.09) \\
  \hline 
  
 \hline
\end{tabular}
\end{center}
\vspace{-10pt}
\captionsetup{justification=centering}
\caption{\small Results at the best test set for our different help models. Each cell shows the mean and standard deviation (std. in parenthesis) for each metric. Gold blocks placed mean is 3.40 and STD is 3.53. All forms of help provided as additional model input in natural language ($\mathtt{M2} - \mathtt{M5}$) improve model performance from the baseline $\mathtt{M1:}$ on both mean distance (lower is better) and mean reward (higher is better), showing how help can be useful for the IGLU task (std. worsens in some cases, but this is due to outliers, see Sec.~\ref{sec:discussion}). Moreover, help is followed a majority of the time by the models, showing that they can successfully incorporate it.}
\vspace{-10pt}
\label{table:results_test_set}
\end{table*}

\subsection{Data}
\label{sec:data}
We use the IGLU Multi-Turn Dataset~\cite{mohanty2023transforming}, which breaks down the complicated IGLU task of building a target structure into steps. We train and evaluate our models at each step. The input to our model is  the most recent Architect instruction and language context (prior instructions), while the output is a sentence describing where blocks should be placed (parsing this output is discussed in Sec.~\ref{sec:task_specific_models}). Data split details: App.~\ref{app:dataset}.

Note that our single-step dialogue-only setup is different from the general IGLU task, which is why we establish our own baselines. Our models are not comparable to the reinforcement learning or clarification question IGLU sub-tasks \cite{kiseleva2022iglu}, as we do not train a first-person 3D RL agent and we ask clarification questions based on confusion to improve final task performance. Thus, we use some different metrics, explained below.

\subsection{Evaluation Metrics} 
Our evaluation framework incorporates four distinct metrics, one of which is used by other IGLU models, while the others are tailored to our unique approach. We evaluate both mean and standard deviation (STD), but prioritize mean, as a higher STD likely results from outliers due to a sub-par baseline model (we discuss this in detail in Sec.~\ref{sec:discussion}).

The first, \textit{IGLU Reward}, determines the invariant intersection between the predicted grid and the target grid~\cite{zholus2022iglu}, which is a primary metric used for evaluation in the IGLU task. (code\footnote{\small \texttt{argmax\_intersection} function: \url{https://github.com/iglu-contest/gridworld/}}). We aim to achieve a high score on this.

The second, \textit{Distance} (Euclidean squared), determines how close on average each model block prediction is to the closest block in the target (lower = better). To account for the difference in \# of blocks placed, we multiply the distance by 1 plus the difference between the \# of blocks predicted and the \# of blocks in the gold grid. If no blocks are predicted, distance is set to a high value of $100$.

The third, \textit{\# Blocks Placed} evaluates how many blocks the model places. This is important as not only do most IGLU instructions require multiple blocks to be placed, but also to make sure the model is outputting valid block dialogue sentences (outputs must be of a certain format to be parsed into coordinates, as discussed in Sec.~\ref{sec:task_specific_models}).

\begin{table}[ht!]
%\small{
\begin{center}
\begin{tabular}{|p{3.4cm}|c|c|c|}
  \hline
  {\textbf{Help Type}} & {\textbf{Train}} & {\textbf{Valid}} & {\textbf{Test}}\\
 \hline
 \small Restrictive & \small 65.88 & \small 66.56 & \small 62.35\\
 \small Corrective & \small 58.12 & \small 55.48 & \small 29.88 \\
 \small Length-based & \small 99.28 & \small 52.12 & \small 40.22 \\
 \small Mistake-Based & \small 98.35 & \small 82.08 & \small 70.40 \\
 \hline
\end{tabular}
\vspace{-5pt}
\end{center}
\caption{\small Accuracy of model self-generated help at training, valdation, and test time.}

\label{table:self_gen_help}
%}
\vspace{-10pt}
\end{table}

The final, \textit{Help Followed}, evaluates how often on average the model correctly follows the help. i.e. placing the block in the correct region (restrictive).

\begin{table*}
\begin{center}
\begin{tabular}{|p{5.2cm}|p{2.4cm}|p{2.1cm}|p{2.1cm}|p{1.8cm}|}
  \hline
  {\textbf{Model}} & {\textbf{ Distance}}
  & {\textbf{Reward}}  & {\textbf{\# Blocks Placed}} & {\textbf{\% Help Followed}} \\

 \hline
 \small $\mathtt{A1:}$ BART Language Model & \small 12.64 (51.75) & \small 1.26 (1.49) & \small 2.56 (2.10) & \small 86.78 (33.86) \\
 \hline 
 \small $\mathtt{A2:}$ Restrictive Help Model Add. Input & \small 10.62 (48.63) & \small 1.38 (1.54) & \small 2.90 (2.34) & \small 81.60 (38.74) \\
 \small $\mathtt{A3:}$ Corrective Help Model Add. Input & \small \textbf{5.10 (9.10)} & \small \textbf{1.74 (1.83)}  & \small \textbf{3.28 (2.47)} & \small 71.98 (44.90) \\
 \small $\mathtt{A4:}$ Length Help Model Add. Input & \small 29.13 (103.22) & \small 0.92 (1.07) &  \small 2.04 (1.50) & \small 46.26 (49.86) \\
 \small $\mathtt{A5:}$ Mistake Help Model Add. Input & \small 11.39 (75.17) & \small 1.67 (1.73) & \small 3.09 (2.51) & \small 95.11 (21.55) \\
  \hline 
  \small $\mathtt{A6:}$ Clarification Questions & \small 13.07 (63.06) & \small 1.29 (1.56) & \small 2.62 (2.07) & \small 67.52 (46.86) \\
  \hline

 \hline
\end{tabular}
\end{center}
\vspace{-5pt}
\captionsetup{justification=centering}
\caption{\small Results at the best test set for our different help models using self-generated help. Gold blocks placed mean is 3.40 and STD is 3.53. Except for length help, which also has low help prediction accuracy, all forms of self-generated help achieve performance improvements over baselines. This shows that even without any human interactions, help can improve performance, as the model learns to predict and then incorporate the help. Further, generating clarification questions based on model confusion, and then providing accurate help in response to the question also increases performance over the baseline.}
\label{table:results_test_set_self_gen}
\vspace{-15pt}
\end{table*}

\subsection{Help Feedback}
\label{sec:help_results}

Tab.~\ref{table:results_test_set} shows the results on the test set. We compare our models to $\mathtt{M1}$, which is our baseline BART Language model from Sec.~\ref{sec:task_specific_models} that achieved Strong IGLU performance and was used as a baseline in the IGLU competition \cite{kiseleva2022iglu}. While we could use a stronger Language Model as a baseline, it would require significantly more compute and resources, which is why we chose BART-base. Further, our focus in this work is developing an interactive process for IGLU-style tasks, and BART-base provides a reasonable baseline.

When incorporating help into BART as an additional language input, we see performance improvements across all help types ($\mathtt{M2} - \mathtt{M5}$), showing that the model can take advantage of all help. Notably, mistake help improves average reward by $\sim$25\%, and corrective help also leads to large improvements. Further, the model follows all help with higher than random accuracy, showing that it can successfully incorporate the help. This shows that help can be a powerful form of human feedback to significantly improve model performance, and a good way for humans to interact with IGLU-style frameworks. Moreover, it is simple to provide, as it can be done in natural language and is based on high-level concepts. We note that in some cases, STD worsens, but this is due to outliers and our weak baseline model, which we explain further in Sec.~\ref{sec:discussion}.

\subsection{Self-Generated Help and Clarification ?'s}
\label{sec:self_gen_help_results}
Tab.~\ref{table:self_gen_help} shows the results of our self-generated help models from Sec.~\ref{sec:self_gen_help_model}. We achieve high performance for restrictive, corrective, and mistake-based help. However, length-based help struggles, as the BART model struggles to accurately quantify the number of blocks to place.

When self-gen help is used at test time in Tab.~\ref{table:results_test_set_self_gen} instead of fully accurate help in Tab.~\ref{table:results_test_set}, we still notice performance improvements in all settings, except length help, \textbf{without human intervention}. Corrective help performs the best, even achieving a higher reward than when it is provided accurately, leading to our \textbf{best performing model} in both mean and STD. We hypothesize that this occurs as whenever the self-generated help is incorrect, it doesn't significantly affect the model's predictions, as the initial prediction was also likely incorrect (note that both help and the initial prediction are coming from the same model). However, when self-gen help is correct, it likely narrows down the model's initial prediction. In contrast, for fully accurate help, some of the help can confuse the model. For example, for restrictive help, if the model doesn't know how to properly search the region provided by the help, the prediction could be much worse, and likely even random in that region. We hypothesize a better baseline model would lead to improvements in both self-gen and accurate help, but due to compute, leave this for future work.

Tab.~\ref{table:results_test_set_self_gen} $\mathtt{A6}$ shows results when the model receives accurate help from clarification questions, once it determines which one it needs (if any) by providing itself all forms of self-generated help, and using it to self-identify confusion. Results show performance improvements over baselines, showing the promise of this approach for the model to accurately identify confusion. However, results are worse than some self-generated help models, as the model can't always identify when it needs help. We leave the investigation of this to future work.

\section{Discussion}
\label{sec:discussion}
In this section, we analyze our IGLU models with help feedback, by asking the following questions:\newline
    (1) \textit{How many regions is best for restrictive help?} \newline
    (2) \textit{How do we do when help is not accurate?} \newline
    (3) \textit{What happens if we vary the help language?} \newline
    (4) \textit{Why does STD worsen sometimes?}

\noindent
\textbf{Restrictive Help -- Number of Regions}
Tab.~\ref{table:ablation_restrictive_regions} shows an ablation study, where we evaluated the number of regions we used for restrictive help on the test set, and chose 8 regions.

\noindent
\textbf{Handling Inaccurate Help}
Help may not always be accurate, especially if provided by humans or someone trying to confuse the model, but the model should be able to adapt. Tab.~\ref{table:self_gen_help} shows the performance of self-generated help that was provided at test time, and it still leads to improvements (Tab.~\ref{table:results_test_set_self_gen}).

\noindent
\textbf{Varying Help Utterances}
To simulate the large language variety of humans providing help (specifically restrictive), we first generate a variety of help by prompting LLM's to write it, and then ask LLM's to determine which region each help corresponds to (details: App.~\ref{appendix:varying_help}). Once the region is known, we can provide help to the models as normal. Results in Tab.~\ref{table:chat_gpt_help_performance} show that LLMs can effectively determine help regions, showing our approach can handle real human help.

\noindent
\textbf{Improving Mean, but sometimes worsening Standard Deviation:} 
While our models always improve mean performance, in some of our experiments (but not all), we see results do not improve on STD. We hypothesize that this occurs due to our weak baseline model, not because our protocols are ineffective. A stronger baseline should lead to more consistent results. In short, whenever STD worsens, it is due to outliers. These are cases where although the help was accurate, the model did not understand the initial IGLU instruction, and thus the help confused it, making the initial prediction worse. For example, for restrictive help, if the BART model can't properly search the region provided because it doesn't understand the initial instruction, it may randomly place the block in the region, worsening the prediction. Thus, these cases that lead to a worse STD are already failure cases, and in fact help does improve performance overall. We discuss this further in Sec.~\ref{app:model_inconsistencies}

\section{Conclusion and Future Work}
\label{sec:summary}
In this paper, we proposed an interactive framework for grounded language understanding tasks, specifically inspired by the IGLU task~\cite{kiseleva2021neurips, kiseleva2022iglu}. Our framework enables humans to interact with AI agents through four distinct forms of help feedback, to provide high-level tips based on concepts relevant for the final task. This high-level help is easy to provide and proves beneficial for the AI agent. Additionally, we proposed a mechanism for the AI agent to autonomously detect confusion and ask clarification questions. To do this, we leveraged help feedback by developing a model to self-generate help, provide it to the agent, and ask a clarification question if confusion is detected. Through this approach, we achieved a fully interactive agent capable of both receiving and providing interactions to humans. Our experiments demonstrated performance improvements in these settings. 

Moving forward, our future work will focus on enhancing the performance of clarification questions, and incorporating more types of help. We are also interested in generalizing our contributions to other domains, including tasks that don't require an agent to navigate a 2D/3D space. 

We believe our approach is directly generalizable to tasks that require an agent to navigate a 3D or 2D space to make decisions (like many robotics tasks). Here, the forms of interactions we proposed and how they are used would not change. For tasks that do not have a 3D/2D space, like summarization, we hypothesize that our framework can still be applicable, by modifying help and keeping the rest of the framework the same. For example, for summarization, the agent must read and summarize certain areas of the document, while ignoring other irrelevant areas, in order to produce a successful summary. Thus, instead of restrictive help restricting the search space of the agent to an area on a 3D grid, restrictive help can restrict the lines of text in a document that the agent has to read. This help would useful to enable the agent to prioritize the relevant parts of the document, leading to a better summary. Similarly, instead of corrective help changing the direction of the grid the agent should search, it can correct the summarization by detailing topics that were missed in the summary. In these ways, our framework can generalize to other tasks, making them end-to-end interactive, and improving performance. Investigating this is part of our future work.

\section{Limitations}
\label{sec:limitations_all}
In this section, we first discuss some limitations of our model and framework (Sec.~\ref{sec:limitations}). Then, we expand with a discussion on ethics as it relates to the deployment of our models (Sec.~\ref{sec:ethics}).

\subsection{Limitations}
\label{sec:limitations}
Our model has been trained on the IGLU \cite{kiseleva2022iglu, kiseleva2021neurips} dataset. Although in the paper we provided results to demonstrate strong performance on this dataset and we hypothesize that our results will generalize to our AI agent instruction following tasks, we have not tested these hypothesis yet, and it is part of our future work. However, we believe our interactive framework of an agent receiving help based on concepts relevant for its task to and also identifying confusion to ask relevant clarifying questions is a general contribution and may be applicable in other scenarios.

Scaling our models to larger settings on larger datasets would likely require more compute, and could impact performance/training time. We trained on a single NVIDIA 12 GB Titan X GPU, and training took a day. Running hyper-parameter search also took a week, to find the best parameters for our Large Language Model.

\subsection{Ethics}
\label{sec:ethics}
 To the best of our knowledge we did not violate any code of ethics through the experiments done in this paper. We reported the details of our experiments both in the main body of the paper and the appendix, including hyperparameter details, training/validation set performance, etc. Moreover, qualitative result we report is an outcome from a machine learning model and does not represent the authors' personal views. 

 Our interactive framework in general should be used to improve the performance of AI agents. However, we understand that some users may use it with malicious intent, such as providing incorrect help feedback to make the agent make a wrong prediction. We showed in the paper, especially in the model self-generated help and discussion sections, that our model can adapt to incorrect human feedback, since the model does not solely rely on human feedback, but also utilizes the knowledge it learns in the training data. However, studying malicious human feedback is an ongoing area of our future work, and users deploying this system should be aware of this possibility.

\bibliography{custom}

\begin{thebibliography}{47}
\expandafter\ifx\csname natexlab\endcsname\relax\def\natexlab#1{#1}\fi

\bibitem[{Akyurek et~al.(2023)Akyurek, Akyurek, Kalyan, Clark, Wijaya, and Tandon}]{akyurek-etal-2023-rl4f}
Afra~Feyza Akyurek, Ekin Akyurek, Ashwin Kalyan, Peter Clark, Derry~Tanti Wijaya, and Niket Tandon. 2023.
\newblock \href {https://doi.org/10.18653/v1/2023.acl-long.427} {{RL}4{F}: Generating natural language feedback with reinforcement learning for repairing model outputs}.
\newblock In \emph{Proceedings of the 61st Annual Meeting of the Association for Computational Linguistics (Volume 1: Long Papers)}, pages 7716--7733, Toronto, Canada. Association for Computational Linguistics.

\bibitem[{Aliannejadi et~al.(2020)Aliannejadi, Kiseleva, Chuklin, Dalton, and Burtsev}]{aliannejadi2020convai3}
Mohammad Aliannejadi, Julia Kiseleva, Aleksandr Chuklin, Jeff Dalton, and Mikhail Burtsev. 2020.
\newblock Convai3: Generating clarifying questions for open-domain dialogue systems (clariq).
\newblock \emph{arXiv preprint arXiv:2009.11352}.

\bibitem[{Aliannejadi et~al.(2021)Aliannejadi, Kiseleva, Chuklin, Dalton, and Burtsev}]{aliannejadi-etal-2021-building}
Mohammad Aliannejadi, Julia Kiseleva, Aleksandr Chuklin, Jeff Dalton, and Mikhail Burtsev. 2021.
\newblock \href {https://doi.org/10.18653/v1/2021.emnlp-main.367} {Building and evaluating open-domain dialogue corpora with clarifying questions}.
\newblock In \emph{Proceedings of the 2021 Conference on Empirical Methods in Natural Language Processing}, pages 4473--4484, Online and Punta Cana, Dominican Republic. Association for Computational Linguistics.

\bibitem[{Arabzadeh et~al.(2022)Arabzadeh, Seifikar, and Clarke}]{arabzadeh2022unsupervised}
Negar Arabzadeh, Mahsa Seifikar, and Charles~LA Clarke. 2022.
\newblock Unsupervised question clarity prediction through retrieved item coherency.
\newblock In \emph{Proceedings of the 31st ACM International Conference on Information \& Knowledge Management}, pages 3811--3816.

\bibitem[{Benotti et~al.(2014)Benotti, Lau, and Villalba}]{benotti2014interpreting}
Luciana Benotti, Tessa Lau, and Mart{\'\i}n Villalba. 2014.
\newblock Interpreting natural language instructions using language, vision, and behavior.
\newblock \emph{ACM Transactions on Interactive Intelligent Systems (TiiS)}, 4(3):1--22.

\bibitem[{Bisk et~al.(2016)Bisk, Yuret, and Marcu}]{bisk2016natural}
Yonatan Bisk, Deniz Yuret, and Daniel Marcu. 2016.
\newblock Natural language communication with robots.
\newblock In \emph{Proceedings of the 2016 Conference of the North American Chapter of the Association for Computational Linguistics: Human Language Technologies}, pages 751--761.

\bibitem[{Borges et~al.(2023)Borges, Tandon, K{\"a}ser, and Bosselut}]{borges2023let}
Beatriz Borges, Niket Tandon, Tanja K{\"a}ser, and Antoine Bosselut. 2023.
\newblock Let me teach you: Pedagogical foundations of feedback for language models.
\newblock \emph{arXiv preprint arXiv:2307.00279}.

\bibitem[{Bu{\ss} and Schlangen(2011)}]{buss2011dium}
Okko Bu{\ss} and David Schlangen. 2011.
\newblock Dium--an incremental dialogue manager that can produce self-corrections.
\newblock \emph{Proceedings of SemDial 2011 (Los Angelogue)}.

\bibitem[{Clark(1996)}]{clark1996using}
Herbert~H Clark. 1996.
\newblock \emph{Using language}.
\newblock Cambridge university press.

\bibitem[{Dalvi et~al.(2022)Dalvi, Tafjord, and Clark}]{dalvi2022towards}
Bhavana Dalvi, Oyvind Tafjord, and Peter Clark. 2022.
\newblock Towards teachable reasoning systems.
\newblock \emph{arXiv preprint arXiv:2204.13074}.

\bibitem[{Elgohary et~al.(2021)Elgohary, Meek, Richardson, Fourney, Ramos, and Awadallah}]{elgohary2021nl}
Ahmed Elgohary, Christopher Meek, Matthew Richardson, Adam Fourney, Gonzalo Ramos, and Ahmed~Hassan Awadallah. 2021.
\newblock Nl-edit: Correcting semantic parse errors through natural language interaction.
\newblock \emph{arXiv preprint arXiv:2103.14540}.

\bibitem[{Fan et~al.(2022)Fan, Wang, Jiang, Mandlekar, Yang, Zhu, Tang, Huang, Zhu, and Anandkumar}]{fanminedojo}
Linxi Fan, Guanzhi Wang, Yunfan Jiang, Ajay Mandlekar, Yuncong Yang, Haoyi Zhu, Andrew Tang, De-An Huang, Yuke Zhu, and Anima Anandkumar. 2022.
\newblock Minedojo: Building open-ended embodied agents with internet-scale knowledge.
\newblock In \emph{Thirty-sixth Conference on Neural Information Processing Systems Datasets and Benchmarks Track}.

\bibitem[{Gluck and Laird(2018)}]{gluck2018interactive}
Kevin~A Gluck and John~E Laird. 2018.
\newblock \emph{Interactive task learning: Humans, robots, and agents acquiring new tasks through natural interactions.}
\newblock The MIT Press.

\bibitem[{Jayannavar et~al.(2020)Jayannavar, Narayan-Chen, and Hockenmaier}]{jayannavar2020learning}
Prashant Jayannavar, Anjali Narayan-Chen, and Julia Hockenmaier. 2020.
\newblock Learning to execute instructions in a minecraft dialogue.
\newblock In \emph{Proceedings of the 58th annual meeting of the association for computational linguistics}, pages 2589--2602.

\bibitem[{Kim et~al.(2021)Kim, Wang, Lee, and Kim}]{kim2021deciding}
Joo-Kyung Kim, Guoyin Wang, Sungjin Lee, and Young-Bum Kim. 2021.
\newblock Deciding whether to ask clarifying questions in large-scale spoken language understanding.
\newblock In \emph{2021 IEEE Automatic Speech Recognition and Understanding Workshop (ASRU)}, pages 869--876. IEEE.

\bibitem[{Kingma and Ba(2014)}]{kingma2014adam}
Diederik~P Kingma and Jimmy Ba. 2014.
\newblock Adam: A method for stochastic optimization.
\newblock \emph{arXiv preprint arXiv:1412.6980}.

\bibitem[{Kiseleva et~al.(2022{\natexlab{a}})Kiseleva, Li, Aliannejadi, Mohanty, ter Hoeve, Burtsev, Skrynnik, Zholus, Panov, Srinet, Szlam, Sun, Hofmann, C{\^o}t{\'e}, Awadallah, Abdrazakov, Churin, Manggala, Naszadi, van~der Meer, and Kim}]{kiseleva2021neurips}
Julia Kiseleva, Ziming Li, Mohammad Aliannejadi, Shrestha Mohanty, Maartje ter Hoeve, Mikhail Burtsev, Alexey Skrynnik, Artem Zholus, Aleksandr Panov, Kavya Srinet, Arthur Szlam, Yuxuan Sun, Katja Hofmann, Marc-Alexandre C{\^o}t{\'e}, Ahmed Awadallah, Linar Abdrazakov, Igor Churin, Putra Manggala, Kata Naszadi, Michiel van~der Meer, and Taewoon Kim. 2022{\natexlab{a}}.
\newblock \href {https://proceedings.mlr.press/v176/kiseleva22a.html} {Interactive grounded language understanding in a collaborative environment: Iglu 2021}.
\newblock In \emph{Proceedings of the NeurIPS 2021 Competitions and Demonstrations Track}, volume 176 of \emph{Proceedings of Machine Learning Research}, pages 146--161. PMLR.

\bibitem[{Kiseleva et~al.(2022{\natexlab{b}})Kiseleva, Skrynnik, Zholus, Mohanty, Arabzadeh, C\^{o}t\'e, Aliannejadi, Teruel, Li, Burtsev, ter Hoeve, Volovikova, Panov, Sun, Srinet, Szlam, Awadallah, Rho, Kwon, Wontae~Nam, Bivort~Haiek, Zhang, Abdrazakov, Qingyam, Zhang, and Guo}]{kiseleva2022iglu}
Julia Kiseleva, Alexey Skrynnik, Artem Zholus, Shrestha Mohanty, Negar Arabzadeh, Marc-Alexandre C\^{o}t\'e, Mohammad Aliannejadi, Milagro Teruel, Ziming Li, Mikhail Burtsev, Maartje ter Hoeve, Zoya Volovikova, Aleksandr Panov, Yuxuan Sun, Kavya Srinet, Arthur Szlam, Ahmed Awadallah, Seungeun Rho, Taehwan Kwon, Daniel Wontae~Nam, Felipe Bivort~Haiek, Edwin Zhang, Linar Abdrazakov, Guo Qingyam, Jason Zhang, and Zhibin Guo. 2022{\natexlab{b}}.
\newblock \href {https://proceedings.mlr.press/v220/kiseleva23a.html} {Interactive grounded language understanding in a collaborative environment: Retrospective on iglu 2022 competition}.
\newblock In \emph{Proceedings of the NeurIPS 2022 Competitions Track}, volume 220 of \emph{Proceedings of Machine Learning Research}, pages 204--216. PMLR.

\bibitem[{Kiseleva et~al.(2016)Kiseleva, Williams, Jiang, Hassan~Awadallah, Crook, Zitouni, and Anastasakos}]{kiseleva2016understanding}
Julia Kiseleva, Kyle Williams, Jiepu Jiang, Ahmed Hassan~Awadallah, Aidan~C Crook, Imed Zitouni, and Tasos Anastasakos. 2016.
\newblock Understanding user satisfaction with intelligent assistants.
\newblock In \emph{Proceedings of the 2016 ACM on conference on human information interaction and retrieval}, pages 121--130.

\bibitem[{Koller et~al.(2010)Koller, Striegnitz, Gargett, Byron, Cassell, Dale, Moore, and Oberlander}]{koller2010report}
Alexander Koller, Kristina Striegnitz, Andrew Gargett, Donna Byron, Justine Cassell, Robert Dale, Johanna~D Moore, and Jon Oberlander. 2010.
\newblock Report on the second nlg challenge on generating instructions in virtual environments (give-2).
\newblock In \emph{Proceedings of the 6th international natural language generation conference}.

\bibitem[{Lewis et~al.(2019)Lewis, Liu, Goyal, Ghazvininejad, Mohamed, Levy, Stoyanov, and Zettlemoyer}]{lewis2019bart}
Mike Lewis, Yinhan Liu, Naman Goyal, Marjan Ghazvininejad, Abdelrahman Mohamed, Omer Levy, Ves Stoyanov, and Luke Zettlemoyer. 2019.
\newblock Bart: Denoising sequence-to-sequence pre-training for natural language generation, translation, and comprehension.
\newblock \emph{arXiv preprint arXiv:1910.13461}.

\bibitem[{Madaan et~al.(2023)Madaan, Tandon, Gupta, Hallinan, Gao, Wiegreffe, Alon, Dziri, Prabhumoye, Yang et~al.}]{madaan2023self}
Aman Madaan, Niket Tandon, Prakhar Gupta, Skyler Hallinan, Luyu Gao, Sarah Wiegreffe, Uri Alon, Nouha Dziri, Shrimai Prabhumoye, Yiming Yang, et~al. 2023.
\newblock Self-refine: Iterative refinement with self-feedback.
\newblock \emph{arXiv preprint arXiv:2303.17651}.

\bibitem[{Madaan et~al.(2021)Madaan, Tandon, Rajagopal, Yang, Clark, Sakaguchi, and Hovy}]{madaan2021improving}
Aman Madaan, Niket Tandon, Dheeraj Rajagopal, Yiming Yang, Peter Clark, Keisuke Sakaguchi, and Ed~Hovy. 2021.
\newblock Improving neural model performance through natural language feedback on their explanations.
\newblock \emph{arXiv preprint arXiv:2104.08765}.

\bibitem[{Manggala and Monz(2023)}]{manggala2023aligning}
Putra Manggala and Christof Monz. 2023.
\newblock Aligning predictive uncertainty with clarification questions in grounded dialog.
\newblock In \emph{Findings of the Association for Computational Linguistics: EMNLP 2023}, pages 14988--14998.

\bibitem[{Mehta and Goldwasser(2019)}]{mehta2019improving}
Nikhil Mehta and Dan Goldwasser. 2019.
\newblock Improving natural language interaction with robots using advice.
\newblock \emph{arXiv preprint arXiv:1905.04655}.

\bibitem[{Milani et~al.(2023)Milani, Kanervisto, Ramanauskas, Schulhoff, Houghton, and Shah}]{milani2023bedd}
Stephanie Milani, Anssi Kanervisto, Karolis Ramanauskas, Sander Schulhoff, Brandon Houghton, and Rohin Shah. 2023.
\newblock Bedd: The minerl basalt evaluation and demonstrations dataset for training and benchmarking agents that solve fuzzy tasks.
\newblock \emph{arXiv preprint arXiv:2312.02405}.

\bibitem[{Mohanty et~al.(2023)Mohanty, Arabzadeh, Kiseleva, Zholus, Teruel, Awadallah, Sun, Srinet, and Szlam}]{mohanty2023transforming}
Shrestha Mohanty, Negar Arabzadeh, Julia Kiseleva, Artem Zholus, Milagro Teruel, Ahmed Awadallah, Yuxuan Sun, Kavya Srinet, and Arthur Szlam. 2023.
\newblock Transforming human-centered ai collaboration: Redefining embodied agents capabilities through interactive grounded language instructions.
\newblock \emph{arXiv preprint arXiv:2305.10783}.

\bibitem[{Mohanty et~al.(2022)Mohanty, Arabzadeh, Teruel, Sun, Zholus, Skrynnik, Burtsev, Srinet, Panov, Szlam et~al.}]{mohanty2022collecting}
Shrestha Mohanty, Negar Arabzadeh, Milagro Teruel, Yuxuan Sun, Artem Zholus, Alexey Skrynnik, Mikhail Burtsev, Kavya Srinet, Aleksandr Panov, Arthur Szlam, et~al. 2022.
\newblock Collecting interactive multi-modal datasets for grounded language understanding.
\newblock \emph{arXiv preprint arXiv:2211.06552}.

\bibitem[{Narayan-Chen et~al.(2017)Narayan-Chen, Graber, Das, Islam, Dan, Natarajan, Doppa, Hockenmaier, Palmer, and Roth}]{narayan2017towards}
Anjali Narayan-Chen, Colin Graber, Mayukh Das, Md~Rakibul Islam, Soham Dan, Sriraam Natarajan, Janardhan~Rao Doppa, Julia Hockenmaier, Martha Palmer, and Dan Roth. 2017.
\newblock Towards problem solving agents that communicate and learn.
\newblock In \emph{Proceedings of the First Workshop on Language Grounding for Robotics}, pages 95--103.

\bibitem[{Narayan-Chen et~al.(2019)Narayan-Chen, Jayannavar, and Hockenmaier}]{narayan-chen-etal-2019-collaborative}
Anjali Narayan-Chen, Prashant Jayannavar, and Julia Hockenmaier. 2019.
\newblock \href {https://doi.org/10.18653/v1/P19-1537} {Collaborative dialogue in {M}inecraft}.
\newblock In \emph{Proceedings of the 57th Annual Meeting of the Association for Computational Linguistics}, pages 5405--5415, Florence, Italy. Association for Computational Linguistics.

\bibitem[{{OpenAI}(2023)}]{chatgpt}
{OpenAI}. 2023.
\newblock {ChatGPT}.
\newblock \url{https://www.openai.com}.

\bibitem[{Padmakumar et~al.(2022)Padmakumar, Thomason, Shrivastava, Lange, Narayan-Chen, Gella, Piramuthu, Tur, and Hakkani-Tur}]{padmakumar2022teach}
Aishwarya Padmakumar, Jesse Thomason, Ayush Shrivastava, Patrick Lange, Anjali Narayan-Chen, Spandana Gella, Robinson Piramuthu, Gokhan Tur, and Dilek Hakkani-Tur. 2022.
\newblock Teach: Task-driven embodied agents that chat.
\newblock In \emph{Proceedings of the AAAI Conference on Artificial Intelligence}, volume~36, pages 2017--2025.

\bibitem[{Paul et~al.(2023)Paul, Ismayilzada, Peyrard, Borges, Bosselut, West, and Faltings}]{paul2023refiner}
Debjit Paul, Mete Ismayilzada, Maxime Peyrard, Beatriz Borges, Antoine Bosselut, Robert West, and Boi Faltings. 2023.
\newblock Refiner: Reasoning feedback on intermediate representations.
\newblock \emph{arXiv preprint arXiv:2304.01904}.

\bibitem[{Raffel et~al.(2020)Raffel, Shazeer, Roberts, Lee, Narang, Matena, Zhou, Li, Liu et~al.}]{raffel2020exploring}
Colin Raffel, Noam Shazeer, Adam Roberts, Katherine Lee, Sharan Narang, Michael Matena, Yanqi Zhou, Wei Li, Peter~J Liu, et~al. 2020.
\newblock Exploring the limits of transfer learning with a unified text-to-text transformer.
\newblock \emph{J. Mach. Learn. Res.}, 21(140):1--67.

\bibitem[{Ren et~al.(2023)Ren, Dixit, Bodrova, Singh, Tu, Brown, Xu, Takayama, Xia, Varley et~al.}]{ren2023robots}
Allen~Z Ren, Anushri Dixit, Alexandra Bodrova, Sumeet Singh, Stephen Tu, Noah Brown, Peng Xu, Leila Takayama, Fei Xia, Jake Varley, et~al. 2023.
\newblock Robots that ask for help: Uncertainty alignment for large language model planners.
\newblock \emph{arXiv preprint arXiv:2307.01928}.

\bibitem[{Ren et~al.(2021)Ren, Xiao, Chang, Huang, Li, Gupta, Chen, and Wang}]{ren2021survey}
Pengzhen Ren, Yun Xiao, Xiaojun Chang, Po-Yao Huang, Zhihui Li, Brij~B Gupta, Xiaojiang Chen, and Xin Wang. 2021.
\newblock A survey of deep active learning.
\newblock \emph{ACM computing surveys (CSUR)}, 54(9):1--40.

\bibitem[{Shah et~al.(2021)Shah, Wild, Wang, Alex, Houghton, Guss, Mohanty, Kanervisto, Milani, Topin et~al.}]{shah2021minerl}
Rohin Shah, Cody Wild, Steven~H Wang, Neel Alex, Brandon Houghton, William Guss, Sharada Mohanty, Anssi Kanervisto, Stephanie Milani, Nicholay Topin, et~al. 2021.
\newblock The minerl basalt competition on learning from human feedback.
\newblock \emph{arXiv preprint arXiv:2107.01969}.

\bibitem[{Shapira et~al.(2021)Shapira, Pasunuru, Ronen, Bansal, Amsterdamer, and Dagan}]{shapira2021extending}
Ori Shapira, Ramakanth Pasunuru, Hadar Ronen, Mohit Bansal, Yael Amsterdamer, and Ido Dagan. 2021.
\newblock Extending multi-document summarization evaluation to the interactive setting.
\newblock In \emph{Proceedings of the 2021 Conference of the North American Chapter of the Association for Computational Linguistics: Human Language Technologies}, pages 657--677.

\bibitem[{Shi et~al.(2022)Shi, Feng, and Lipani}]{shi2022learning}
Zhengxiang Shi, Yue Feng, and Aldo Lipani. 2022.
\newblock Learning to execute actions or ask clarification questions.
\newblock In \emph{Findings of the Association for Computational Linguistics: NAACL 2022}, pages 2060--2070.

\bibitem[{Shridhar et~al.(2020)Shridhar, Thomason, Gordon, Bisk, Han, Mottaghi, Zettlemoyer, and Fox}]{shridhar2020alfred}
Mohit Shridhar, Jesse Thomason, Daniel Gordon, Yonatan Bisk, Winson Han, Roozbeh Mottaghi, Luke Zettlemoyer, and Dieter Fox. 2020.
\newblock Alfred: A benchmark for interpreting grounded instructions for everyday tasks.
\newblock In \emph{Proceedings of the IEEE/CVF conference on computer vision and pattern recognition}, pages 10740--10749.

\bibitem[{Skrynnik et~al.(2022)Skrynnik, Volovikova, C{\^o}t{\'e}, Voronov, Zholus, Arabzadeh, Mohanty, Teruel, Awadallah, Panov et~al.}]{skrynnik2022learning}
Alexey Skrynnik, Zoya Volovikova, Marc-Alexandre C{\^o}t{\'e}, Anton Voronov, Artem Zholus, Negar Arabzadeh, Shrestha Mohanty, Milagro Teruel, Ahmed Awadallah, Aleksandr Panov, et~al. 2022.
\newblock Learning to solve voxel building embodied tasks from pixels and natural language instructions.
\newblock \emph{arXiv preprint arXiv:2211.00688}.

\bibitem[{Tandon et~al.(2022)Tandon, Madaan, Clark, and Yang}]{tandon2022learning}
Niket Tandon, Aman Madaan, Peter Clark, and Yiming Yang. 2022.
\newblock Learning to repair: Repairing model output errors after deployment using a dynamic memory of feedback.
\newblock \emph{NAACL Findings.(to appear)}.

\bibitem[{Wang et~al.(2023)Wang, Xie, Jiang, Mandlekar, Xiao, Zhu, Fan, and Anandkumar}]{wang2023voyager}
Guanzhi Wang, Yuqi Xie, Yunfan Jiang, Ajay Mandlekar, Chaowei Xiao, Yuke Zhu, Linxi Fan, and Anima Anandkumar. 2023.
\newblock Voyager: An open-ended embodied agent with large language models.
\newblock \emph{arXiv preprint arXiv:2305.16291}.

\bibitem[{White et~al.(2021)White, Poesia, Hawkins, Sadigh, and Goodman}]{white2021open}
Julia White, Gabriel Poesia, Robert Hawkins, Dorsa Sadigh, and Noah Goodman. 2021.
\newblock Open-domain clarification question generation without question examples.
\newblock In \emph{Proceedings of the 2021 Conference on Empirical Methods in Natural Language Processing}, pages 563--570.

\bibitem[{Winograd(1972)}]{winograd1972understanding}
Terry Winograd. 1972.
\newblock Understanding natural language.
\newblock \emph{Cognitive psychology}, 3(1):1--191.

\bibitem[{Wolf et~al.(2020)Wolf, Debut, Sanh, Chaumond, Delangue, Moi, Cistac, Rault, Louf, Funtowicz, Davison, Shleifer, von Platen, Ma, Jernite, Plu, Xu, Scao, Gugger, Drame, Lhoest, and Rush}]{wolf-etal-2020-transformers}
Thomas Wolf, Lysandre Debut, Victor Sanh, Julien Chaumond, Clement Delangue, Anthony Moi, Pierric Cistac, Tim Rault, Rémi Louf, Morgan Funtowicz, Joe Davison, Sam Shleifer, Patrick von Platen, Clara Ma, Yacine Jernite, Julien Plu, Canwen Xu, Teven~Le Scao, Sylvain Gugger, Mariama Drame, Quentin Lhoest, and Alexander~M. Rush. 2020.
\newblock \href {https://www.aclweb.org/anthology/2020.emnlp-demos.6} {Transformers: State-of-the-art natural language processing}.
\newblock In \emph{Proceedings of the 2020 Conference on Empirical Methods in Natural Language Processing: System Demonstrations}, pages 38--45, Online. Association for Computational Linguistics.

\bibitem[{Zholus et~al.(2022)Zholus, Skrynnik, Mohanty, Volovikova, Kiseleva, Szlam, Cot{\'e}, and Panov}]{zholus2022iglu}
Artem Zholus, Alexey Skrynnik, Shrestha Mohanty, Zoya Volovikova, Julia Kiseleva, Artur Szlam, Marc-Alexandre Cot{\'e}, and Aleksandr~I Panov. 2022.
\newblock Iglu gridworld: Simple and fast environment for embodied dialog agents.
\newblock \emph{arXiv preprint arXiv:2206.00142}.

\end{thebibliography}

\newpage
\appendix

\section{Additional Related Works}
\label{appendix:additional_related_works}
\textbf{User-Feedback} As interactive grounded language understanding tasks like IGLU are very challenging, many works have looked at how humans can interact with agents to provide feedback. \cite{benotti2014interpreting} allow humans to rephrase their instructions in feedback. However, on more challenging tasks like IGLU, this new instruction may still be complex enough that the model won't understand it and thus likely won't help the model generalize/learn better. Active learning mechanisms \cite{ren2021survey} show how users can interact with the agent during training, and normally this involves having the agent ask questions when it needs help. We experimented with this as well in our setup, where help is used to identify confusion, enabling the agent to ask clarification questions. \citeauthor{elgohary2021nl} learn to apply user-provided syntactic edit operations. \citeauthor{buss2011dium} show how dialogue models can propose self-corrections, whereas we show how grounded language learning systems can do this, specifically ones that directly help their task. Other works \cite{madaan2021improving, tandon2022learning, dalvi2022towards} show how user-feedback can be used to correct/improve LLMs, even being saved in memory. More recent works \cite{madaan2023self, paul2023refiner} use LLMs to generate the feedback/reasoning steps. Even more recently, \citeauthor{borges2023let} design a general framework, FELT, for how LLM-feedback can be provided, by training a model to provide it. In the future, these works can be combined with our framework, where help is provided via a language model, that is improved using reinforcement learning.

\section{Help Details}
\label{appendix:help_examples}
In this section, we provide detailed examples of the help types discussed in Sec~\ref{sec:subsection_help_types}. Help is generated based on gold data, or in the model self-generated case based on model predictions. Based on these coordinates (either gold or predicted), we can generate the help and fill it into a pre-defined slot based on each help type.

\subsection{Restrictive}
For Restrictive Help, we divide the center region (from -0.5 to 0.5 in the x and y directions) into an equal number of regions (either 4 or 8, depending on the model). Then, we divide the rest of the grid into 4 regions, also in the x and y directions. For example, a coordinate with (x, y) location (0.8, 0.8) is in the `upper left not in the center' region, while a coordinate (0.2, 0.2) is in the `upper left in the center' region. These regions are then filled into the slots in the sentence `Place the block in the \_ region', to form the final \textit{help} sentence: `Place the block in upper left not in the center region'. As we have 8 total regions, the different region descriptions we use are: "upper right", "upper left", "lower left", "lower right", "upper upper right", "upper upper left", "lower lower left", and "lower lower right"

\subsection{Length-Based}
Length-based help tells the model how many blocks to place. For example, if 3 blocks must be placed, then the help is `You should place 3 blocks'. To generate the help utterance, we slot fill the sentence `You should place \_ blocks` with a number representing the number of blocks to place.

\subsection{Corrective}
Corrective help tells the model what direction to adjust its' predictions in. For example, if the model predicted a (x, y) coordinate of (0.5, 0.5) and the true block location was (0.8, 0.5), then the model should place the block more to the right, based on the x coordinate. Thus, the help would be be: `Place the block more to the right'. To generate the help utterance, we slot fill the sentence `Place the block more to the \_ ` with either "left', "right", "up", or "down" (depending on the direction to adjust).

\subsection{Mistake-Based}
Mistake-based help tells the model how many blocks it placed incorrectly. For example, if the model placed 3 blocks and 2 were placed incorrectly, the help would be: `You placed 2 blocks incorrectly'. To generate the help utterance, we slot fill the sentence `You placed \_ blocks incorrectly` with a number corresponding to how many blocks were placed incorrectly.

\section{Implementation Details}
We implement our models using the PyTorch Framework\footnote{\url{https://pytorch.org/}} and use the Transformers package \cite{wolf-etal-2020-transformers} for our Transformer implementations. We use the Facebook BART-Base model everywhere that we use a Transformer Language Model \cite{lewis2019bart}. We train our end-to-end model with a learning rate of 1e-4 and the Adam optimizer \cite{kingma2014adam}. Our self-generated models use a learning rate of 1e-6 and the classification layer is not pre-loaded. We trained all our models using a 12GB TITAN XP GPU card. Training the self-generated model took approximately 5 hours, whereas training the end-to-end models took anywhere from 1-2 days. We mentioned the details of our dataset in Sec.~\ref{sec:data}.

\section{Dataset Details}
\label{app:dataset}
We use the public IGLU MultiTurn Dataset\footnote{\url{https://gitlab.aicrowd.com/aicrowd/challenges/iglu-challenge-2022/iglu-2022-rl-mhb-baseline/-/tree/master/nlp_training}}. The dataset breaks down the complicated IGLU task of building a reference structure into steps, and we train and evaluate our models on each step. Thus, the input to our model is  the most recent Architect instruction and language context (prior Builder/Architect instructions), while the output is a sentence describing where blocks should be placed (if any; parsing this output is discussed in Sec~\ref{sec:task_specific_models}). Training details:~\ref{app:dataset}. We use the \verb|train_data_augmented_part1.json| file for training, and the \verb|val_data.json| file for testing. We have 8,736 training samples, 11,283 validation, and 1,238 test. When evaluating the confusion/clarification question models, we use 50\% of the training/dev data to learn the self-generated help models, and then generate help for the validation/test sets, using gold at train time. For fair comparison, the test sets in all settings are the same.

\section{Ablation Study for Restrictive Help}
\label{appendix:ablation_restrictive}
In Tab.~\ref{table:ablation_restrictive_regions}, we show an ablation study for restrictive help, evaluated on a smaller dataset. It is clear that restrictive help with 8 regions leads to the best performance, which is why we use it.

\begin{table*}
\begin{center}
\begin{tabular}{|p{4.5cm}|p{2.4cm}|p{2.1cm}|p{2.1cm}|p{2.1cm}|}
  \hline
  {\textbf{Model}} & {\textbf{ Distance}}
  & {\textbf{Reward}}  & {\textbf{\# Blocks Placed}} & {\textbf{\% Help Followed}} \\

 \hline 
 Restrictive Help 4 Regions & 23.06 (37.67) & 0.48 (0.73) & 2.32 (1.53) & 69.56 (46.01) \\
 Restrictive Help 8 Regions & 23.03 (27.71) & 0.54 (0.76) & 2.54 (0.70) & 65.22 (47.62) \\
 Restrictive Help 16 Regions & 37.69 (108.12) & 0.42 (0.63) & 2.60 (1.15) & 62.64 (48.37) \\
  \hline 
\end{tabular}
\end{center}
\vspace{-5pt}
\captionsetup{justification=centering}
\caption{Ablation Study: Test Set Results for different number of regions for restrictive help. We find that 8 regions provides the best performance. Gold blocks placed mean is 3.40 and STD is 3.53.}
\label{table:ablation_restrictive_regions}
\end{table*}

\section{Discussion Cont.: Varying Help Utterances}
\label{appendix:varying_help}
In the main experiments of the paper, the help we used was generated by slot-filling to create synthetic utterances, as discussed in Sec.~\ref{sec:slot_filling_help}. However, when real humans provide help, they are likely to provide it via a wide variety of language, not just several pre-defined slots. In this section, we simulate these settings, by first generating large amounts of help utterances that have a variety of language, and then using them as help in our final IGLU Task Model.

\begin{figure*}[t!]
  \centering
  \includegraphics[scale=0.5]{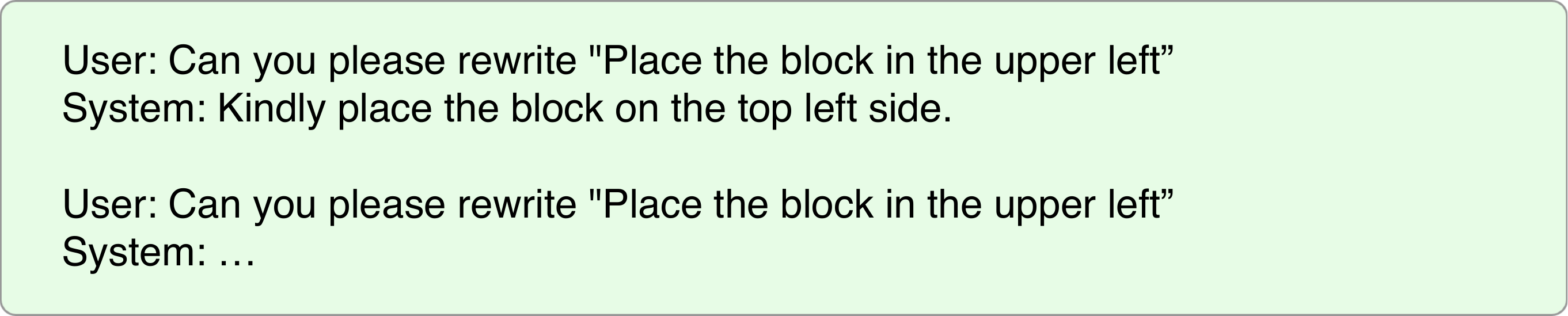}
\caption{An example of the ChatGPT interaction to rewrite utterances. The user asks the system to rewrite help utterances, in this case for the ``upper left'' region. ChatGPT then does it (shown by the ``System'' response). If an utterance is repeated, it is discarded. Finally, all rewrites are manually inspected by humans to make sure they are valid and not conflicting with other regions (such as ``upmost left'' in this case).}
\label{fig:chat_gpt_rewrite}
\end{figure*}

As collecting a large amount of human help interactions is not cost efficient, we simulate these settings, focusing on restrictive help. To get a variety of help utterances, we prompt a strong language generation Large Language Model (LLM), ChatGPT \cite{chatgpt}, to generate them. In the prompt, seen in Fig.~\ref{fig:chat_gpt_rewrite}, we ask the LLM to rewrite utterances in a different way. We manually inspect the outputs to discard duplicates and ensure validity, and keep the rest. 

Once we have a large amount of restrictive help (25 utterances for each region), each written in a different way, we aim to use them in the final IGLU Task Model, as different ways humans can provide help. However, instead of having the IGLU Task Model determine which region each help utterance corresponds to, which could be difficult, we use LLM's to do it. For this, we few-shot prompt ChatGPT, to output the region corresponding to the utterance. Once the region is known, we can feed it directly to the IGLU task model, such as by using the same slot-filling generated utterances from Sec.~\ref{sec:slot_filling_help}, but now generated using the predicted region. Then, the rest of our setup would be identical as before, except now our IGLU Task model can use a wide variety of language as help.

\begin{figure*}[t!]
  \centering
  \includegraphics[scale=0.5]{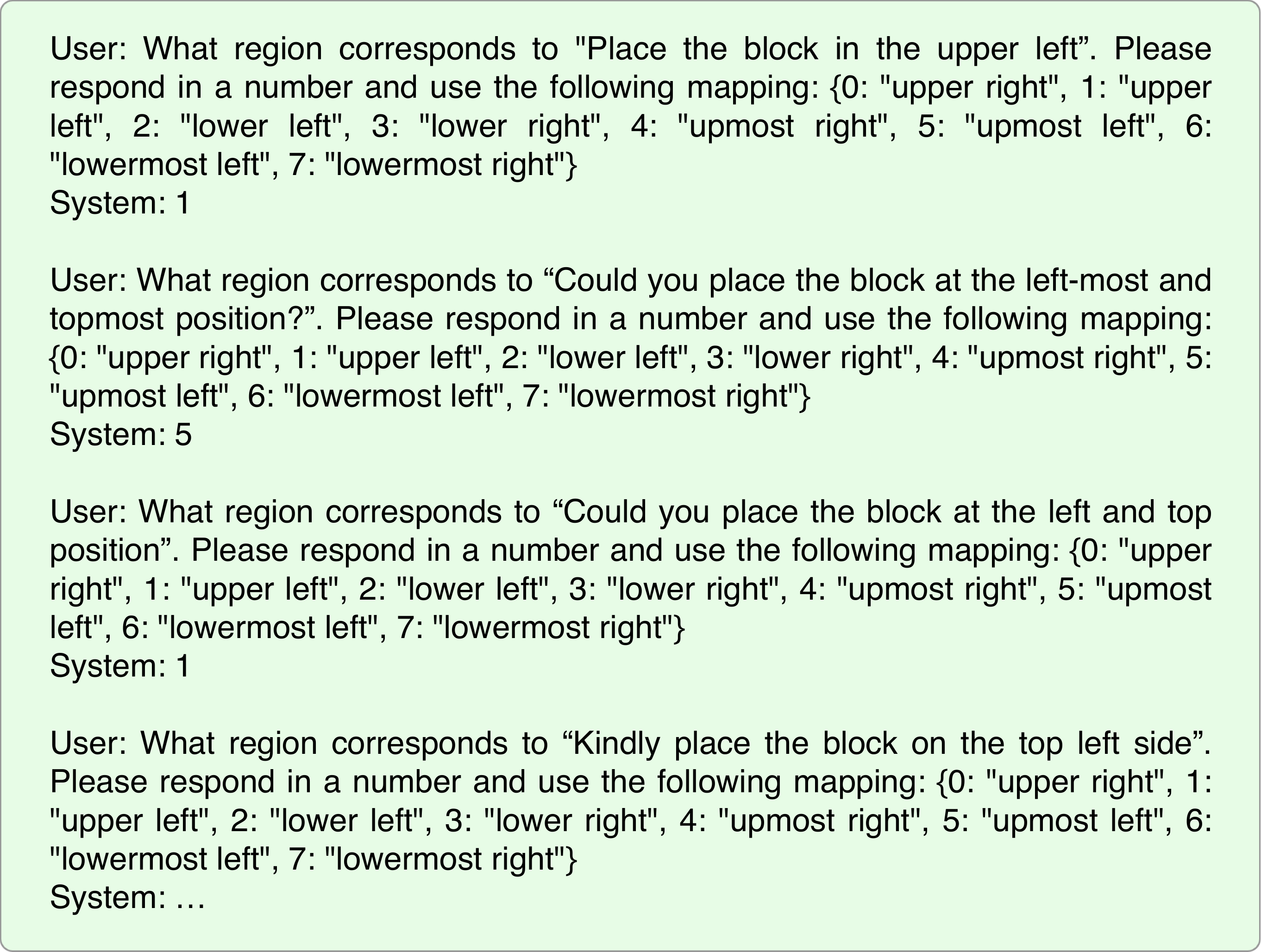}
\caption{An example of the ChatGPT prompt to classify help utterances into a region. ChatGPT is prompted with examples of a region, for the ``upper left'' and ``upmost left'' region, as these could be confusing. It then must output the correct region. The figure shows three few-shot training examples, and then ChatGPT makes a prediction on the last one, shown by ``....''. We use the same ``upper left'' centered few-shot examples for other regions as well, and ChatGPT can generalize.}
\label{fig:chat_gpt_classify}
\end{figure*}

An example of the few-shot training examples and the ChatGPT model output is seen in Fig.~\ref{fig:chat_gpt_classify}. Tab.~\ref{table:chat_gpt_help_performance} shows the results, and we can see that ChatGPT is able to well determine the regions from a variety of help utterances. While we do not evaluate the ChatGPT predictions end-to-end in our IGLU Task Model and instead leave it for future work, we do not expect significant performance changes, given the high performance of ChatGPT to determine the regions correctly.

We believe that these initial results show that our system can handle actual human help, which can have a large amount of language variety. By using ChatGPT to determine which region corresponds to the help and then creating the help utterances for the IGLU Task model using that, we are able to handle the large language variety humans may use when providing help.

\begin{table}[h!]
%\small{
\begin{center}
\begin{tabular}{|p{3.4cm}|}
  \hline
  {\textbf{Region}}\\
 \hline
 %2 & 63\% & 65\% \\
 \small Upper Right \\

 \hline
\end{tabular}
\end{center}
\caption{\small Examples of help utterances generated by ChatGPT when asked to rewrite: ``Place the block in the upper left''. We can see that there is a large variety in the language of the help, similar to how humans would provide the same help with a large amount of language.}
\label{table:chat_gpt_outputs}
%}
\end{table}

\begin{table}[h!]
%\small{
\begin{center}
\begin{tabular}{|p{3.4cm}|c|}
  \hline
  {\textbf{Region}} & {\textbf{Accuracy}}\\
 \hline
 %2 & 63\% & 65\% \\
 \small Upper Right & \small  \\
 \small Upmost Right & \small 85.00 \\
 \small Upper Left & \small 95.45 \\
 \small Upmost Left & \small 93.75 \\
 \small Lower Left & \small 94.44 \\
 \small Lowermost Left & \small 85.71 \\
 \small Lower Right & \small 82.60 \\
 \small Lowermost Right & \small 81.25 \\

 \hline
\end{tabular}
\end{center}
\caption{\small Accuracy of ChatGPT few-shot predicting the correct region for each help utterance, based on 25 utterances. Results show that this is a fairly simple task for ChatGPT, achieving high accuracy for all regions. Thus, we hypothesize our models can handle a variety of language in the help utterances.}
\label{table:chat_gpt_help_performance}
%}
\end{table}

\section{Real World Application of Help}
\label{appendix:real_world_help}
In this section, we discuss a potential real-world application of our help system, enabling humans to communicate with AI agents for tasks like IGLU in natural language.

In this paper, we simulated the help by slot filling pre-defined utterances. However, in the real-world, humans can provide help in a variety of language. To handle this, we first note that each form of help is constrained in some way, i.e. has a limited number of options for the types of help that can be given. For example, restrictive help can has 8 regions, length-based help has a maximum of 8 blocks that can be placed, corrective help has 4 directions to move, and mistake-based help has up to 8 number of blocks that can be placed incorrectly. Thus, every human help utterance must be mapped to one of the options. We hypothesize that this can be done using a few-shot prompted Large Language Model (LLM), where the model is trained for a classification problem. For example, it could be trained to first identify which form of help the human is providing, i.e. restrictive, and then which version of restrictive help, i.e. which region the block should be placed in. This would allow converting a varying language help utterance into one of our "slot-filled" help utterances, and then our framework could be used as normal.

Further, in this paper we only experimented with a single-step dialogue only IGLU setup, but it is possible that IGLU be solved with a different setup, like a Reinforcement Learning (RL) agent. In this case, our help can be provided as an additional input to the RL agent model via a Language Model component, and then everything can be used as normal.

\section{Discussion: Model Inconsistencies}
\label{app:model_inconsistencies}
Our primary novel contribution in this work is our methods for enabling fully interactive systems for challenging grounded language understanding tasks like IGLU, something which is often looked over in today's research. Our experimental results show that our ideas are beneficial. Notably, our best model sees significant performance improvements over our baseline. Table 2, row A3, shows a large performance improvement on distance and number of blocks placed. For example, mean distance (lower is better) improves from the baseline of 12.64 to 5.10 and STD distance improves from 51.73 to 9.10.

In some of our other experiments, while our models always offer performance improvements, results may not improve significantly, particularly on STD. We hypothesize that this happens due to our baseline model not being strong enough on certain examples, not because our protocols are ineffective. A stronger baseline should lead to better results. Unfortunately, due to lack of compute resources, in this paper we could not use a stronger Language Model than BART as a baseline, but we leave the investigation of this to future work.

As a case study, let us look at a test example where the baseline model cannot come close to the correct prediction. In this case, even an accurate human interaction cannot help the model perform better, as humans only aim to help the model, not solve the final task. For example, when using corrective help, if humans tell the model to adjust it's prediction left and the initial prediction is already significantly wrong, the help is likely to not assist and might even make the prediction worse, such as the model going left by a significant amount.

Now, let us see additional evidence of improvements, first looking at all our help models. We see that mean value almost always improves, but in some cases STD worsens. The improving mean shows that in cases where the model can appropriately understand the initial example and thus take advantage of the help, the help improves performance significantly, even if help is self-generated. However, in the cases where the model cannot solve the initial example, help can make the prediction worse (as explained above), leading to a worse STD. Thus, overall, human interactions actually improve the model, only hurting it on examples where it was wrong anyways (thus a worse STD).

Now, let us look at Self-Generated (Table~\ref{table:results_test_set_self_gen}) vs 100\% Accurate Help (Table~\ref{table:results_test_set}) results. We see self-generated corrective help performs better than 100\% accurate corrective help. Why is this? Well, when using self-generated help, the model will predict help accurately for examples it can already solve and for borderline examples. Then, when using the help on these examples, either existing predictions are reinforced, or borderline predictions are corrected, leading to improvements. In the cases the model can't solve at all, it will likely still predict help, but incorrectly. However, it won't be incorrect enough to dramatically change the prediction, since the model's fundamental understanding of the example hasn't changed. Thus, the STD doesn't worsen. In contrast, accurate help may tell the model to significantly change its prediction, confusing it and leading to worse results.

The above shows how human interactions via help does improve our models.

\end{document}